\documentclass{article}

\usepackage[dblblindworkshop, final]{neurips_2026}

\workshoptitle{The 6th Workshop on Mathematical Reasoning and AI}

\usepackage[utf8]{inputenc}
\usepackage[T1]{fontenc}
\usepackage{hyperref}
\usepackage{url}
\usepackage{booktabs}
\usepackage{amsmath}
\usepackage{amsfonts}
\usepackage{amssymb}
\usepackage{graphicx}
\usepackage{tabularx}
\usepackage{nicefrac}
\usepackage{microtype}
\usepackage{xcolor}
\usepackage{enumitem}
\usepackage{capt-of}
\usepackage{listings}
\usepackage{pifont}
\usepackage{tikz}
\usetikzlibrary{arrows.meta}

\newcommand{\CC}{\mathbb{C}}
\newcommand{\QQ}{\mathbb{Q}}

\title{Agentic Computational Algebra}

\author{%
  Ryan A. Anderson\thanks{Work done during an internship at AWS.} \\
  UCLA \\
  \And
  Matthew Trager\thanks{Correspondence: \href{mailto:mttrager@amazon.com}{\texttt{mttrager@amazon.com}}.} \\
  AWS \\
  \And
  Ben Bowman \\
  AWS \\
  \AND
  Wei Xia \\
  AWS \\
  \And
  Stefano Soatto \\
  AWS \\
}

\begin{document}

\maketitle

\begin{abstract}
We explore the use of AI agents for solving computational problems in algebra. For the problems we consider, the main challenge is not correctness but \emph{time}: general-purpose algorithms are guaranteed to provide correct solutions but can be prohibitively slow, while reformulations that exploit special structure often allow solving a problem much faster. We propose to use this setting to evaluate how effectively agents discover and exploit structure within and across tasks to reduce computational costs. More broadly, computational algebra supplies infinitely many problems with independently verifiable answers and systematically controllable structure and difficulty. To illustrate these possibilities, we use benchmark families of systems with controlled shared structure and evaluate agents in three settings: solving independent instances, solving online streams of related instances using persistent memory, and compiling a fast reusable solver for a known problem family.
Our experiments show that (i) agents can extend the range of
systems solvable within a fixed budget, although reasoning
overhead can sometimes outweigh computational savings; (ii) memory can reduce the cost of repeated solving but may preserve suboptimal
strategies; (iii) compilation produces faster reusable solvers
but at substantial upfront cost. Overall, computational algebra offers concrete, measurable targets for improving how agents discover and reuse structure.
\end{abstract}

\section{Introduction}

Many questions in the sciences come down to systems of polynomial equations and inequalities: in robotics, configuring a robot arm to reach a given point \citep{Wampler2011NumericalAG}; in data analysis, recovering a low-rank tensor from linear measurements \citep{tensor-recovery2026}; in economics, computing mixed-strategy Nash equilibria of finite games \citep{nash2026}. Tools from computational algebra---including Gr\"obner bases, resultants, real quantifier elimination, and certified homotopy continuation---provide exact or certified methods for solving such systems~\citep{coxlittleoshea,sturmfels2002}. In practice, however, these approaches are computationally very expensive, and solving even moderately sized systems can be prohibitively costly. Moreover, the same algebraic problem can admit many valid reformulations and solving strategies, whose execution costs can differ by orders of magnitude. Expert judgment is therefore often needed to choose an effective representation and method.

In this paper, we explore whether AI agents can help with these choices by recognizing hidden structure, reformulating problems, and reusing computations across related instances. For the problems we consider, general-purpose algorithms provide a guaranteed route to a correct solution, but an agent can sometimes discover a faster specialized approach. In such cases, the time saved by a faster solution method must outweigh the agent's cost of discovering it; however, savings can accumulate if the method is reused across related instances.  To explore these tradeoffs, we evaluate independent instances, online streams with persistent memory, and compilation of reusable solvers.

More broadly, computational algebra offers a rich domain for studying how agents can exploit structure within and across tasks to reduce costs, a core aspect of general `agentic learning.' Algebraic constructions allow us to generate problem families with known opportunities for simplification or reusable computation, vary their difficulty, and change how structure is presented to a solver. Our experiments illustrate these properties in the three settings above, and provide initial evidence that agents can find approaches that lead to speedups compared to direct solver baselines. However, they also show that reasoning and memory overhead can prevent solver improvements from translating into end-to-end gains, and that agents often remain substantially slower than baselines that exploit known structure.

\paragraph{Related work.} Language models have been used to search for mathematical constructions \citep{funsearch2024} and to solve olympiad geometry problems with a symbolic engine in the loop \citep{alphageometry2024}. Automated research systems such as AlphaEvolve~\citep{alphaevolve2025} and autoresearch~\citep{karpathy2026autoresearch} use evaluator feedback to iteratively improve programs, an approach similar to our compilation regime. The compilation setup also connects to the problem of generating fast specialized polynomial solvers, developed extensively for minimal problems in computer vision~\citep{Kukelova2008AutomaticGO}. Machine learning has also been applied to algebraic tasks such as selecting monomial orderings \citep{bunch2026learningfastmonomialorders} and selecting S-pairs~\citep{Peifer2020LearningSS} in Gr\"obner basis computation.  Persistent agent memory has been studied through self-reflection and reusable skill libraries \citep{reflexion2023,voyager2023}. Our emphasis on computational cost is related to recent work on agents as universal solvers \citep{achille2026agents} and to the evaluation of Las Vegas algorithms \citep{LUBY1993173,hoos2013evaluatinglasvegasalgorithms}.

\section{Setup}
\label{sec:setup}

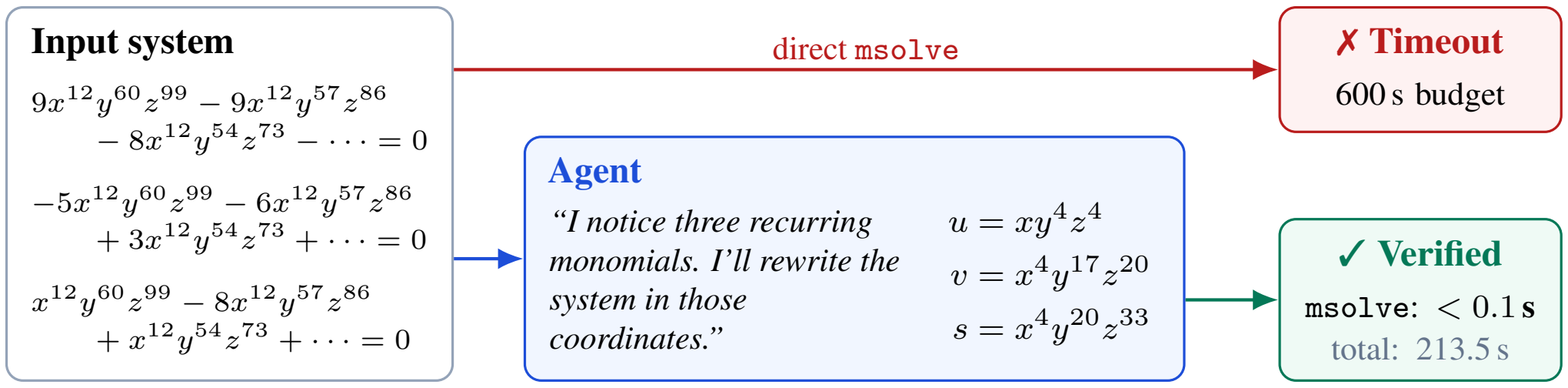
\begin{figure}[t]
\centering
\definecolor{tsLine}{HTML}{94A3B8}
\definecolor{tsFail}{HTML}{B91C1C}
\definecolor{tsFailFill}{HTML}{FEF2F2}
\definecolor{tsAgent}{HTML}{1D4ED8}
\definecolor{tsAgentFill}{HTML}{F1F6FF}
\definecolor{tsPass}{HTML}{047857}
\definecolor{tsPassFill}{HTML}{EFFAF5}
\definecolor{tsMuted}{HTML}{64748B}
\begin{tikzpicture}[
    font=\footnotesize,
    box/.style={draw, rounded corners=4pt, line width=0.6pt, inner xsep=6pt, inner ysep=5pt},
    flow/.style={-{Latex[length=2.2mm, width=1.8mm]}, line width=0.8pt},
    edgelabel/.style={font=\footnotesize, inner sep=2pt},
  ]
  \node[box, draw=tsLine, fill=white, align=left, anchor=north west]
    (sys) at (0, 0) {%
      {\small\bfseries Input system}\\[5pt]
      \renewcommand{\arraystretch}{1.15}%
      \scriptsize
      $\begin{array}{@{}l@{}}
        9x^{12}y^{60}z^{99} - 9x^{12}y^{57}z^{86}\\
        \qquad{}- 8x^{12}y^{54}z^{73} - \cdots = 0\\[6pt]
        -5x^{12}y^{60}z^{99} - 6x^{12}y^{57}z^{86}\\
        \qquad{}+ 3x^{12}y^{54}z^{73} + \cdots = 0\\[6pt]
        x^{12}y^{60}z^{99} - 8x^{12}y^{57}z^{86}\\
        \qquad{}+ x^{12}y^{54}z^{73} + \cdots = 0
      \end{array}$};

  \node[box, draw=tsAgent, fill=tsAgentFill, align=flush left, text width=5.25cm,
        anchor=south west]
    (agent) at ([xshift=6mm]sys.south east) {%
      {\small\bfseries\color{tsAgent}Agent}\\[2pt]
      \begin{tabular}{@{}p{3.15cm}@{\hspace{8pt}}l@{}}
        \raggedright\microtypesetup{protrusion=false}\textit{``I notice three recurring monomials. I'll rewrite
        the system in those coordinates.''}
        &
        $\begin{aligned}[t]
          u &= xy^{4}z^{4}\\
          v &= x^{4}y^{17}z^{20}\\
          s &= x^{4}y^{20}z^{33}
        \end{aligned}$
      \end{tabular}};

  \node[box, draw=tsPass, fill=tsPassFill, align=center, text width=2.0cm,
        anchor=south west]
    (pass) at ([xshift=8mm]agent.south east) {%
      {\small\bfseries\color{tsPass}\ding{51}\ Verified}\\[3pt]
      \texttt{msolve}: \textbf{$<0.1$\,s}\\[1pt]
      \textcolor{tsMuted}{total: 213.5\,s}};
  \node[box, draw=tsFail, fill=tsFailFill, align=center, text width=2.0cm,
        anchor=north]
    (fail) at (pass.north |- sys.north) {%
      {\small\bfseries\color{tsFail}\ding{55}\ Timeout}\\[3pt]
      600\,s budget};

  \draw[flow, tsFail] (sys.east |- fail) -- node[edgelabel, above] {direct \texttt{msolve}} (fail.west);
  \draw[flow, tsAgent] (sys.east |- agent) -- (agent.west);
  \draw[flow, tsPass] (agent.east |- pass) -- (pass.west);
\end{tikzpicture}
\caption{Hidden-monomial example: the agent reduces degree-171 equations to three cubics and returns all 27 solutions, while direct \texttt{msolve} times out. Agent output is paraphrased.}
\label{fig:teaser-monomial}
\end{figure}

\paragraph{Problems and solvers.} A \emph{problem} consists of an input and an ``output contract,'' specified operationally by an independent verification procedure. A \emph{solver} takes an input and attempts to produce a candidate answer that passes the verification. While this setting is general, in this work we focus on solving zero-dimensional systems of polynomial equations, i.e., systems that admit finitely many complex solutions. Unless stated otherwise, we ask for all solutions over the complex numbers in exact form via rational univariate representations (RURs~\citep{Rouillier1999-dg}); see Appendix~\ref{app:domain}.

\paragraph{General-purpose solvers.} Zero-dimensional polynomial systems with rational coefficients can always be solved exactly and systematically. A standard class of approaches relies on computing a Gr\"obner basis: a set of polynomials with the same solution set as the original equations that enables convenient computations using linear algebra, including the construction of an RUR of the solutions. Such a basis can be computed by iteratively applying algebraic reductions, but this is often computationally very expensive.  Solvers such as \texttt{msolve} implement this pipeline~\citep{msolve2021}.
Gr\"obner-basis computations are also available in computer algebra systems such as \texttt{Singular}~\citep{singular}, \texttt{Macaulay2}~\citep{M2}, and \texttt{SageMath}~\citep{sagemath}.

\paragraph{Agentic solvers.} An \emph{agentic solver} uses a language model to organize computation with tools, optionally with a persistent memory for retaining useful strategies or computations across problem instances. In our setting, the agent can inspect the input equations, propose and test reformulations, invoke general-purpose solvers, and assemble an answer that satisfies the output contract. This is illustrated in Figure~\ref{fig:teaser-monomial}.

\paragraph{Problem families.} A \emph{problem family} is a collection of problems defined by a shared parameterized construction. For polynomial systems, the parameters can control features such as size, degree, and coefficients, allowing systematic variation of difficulty and shared structure. Our benchmarks use families with known opportunities for efficient solving and reusable computation. The concrete constructions are described in Appendix~\ref{app:families}.

\paragraph{Evaluation regimes.}
We compare agentic solvers and general-purpose solvers in terms of the computational cost (time) to produce a valid answer. For agents, this includes model reasoning, tool execution, and any memory maintenance, while independent evaluation of the output is excluded. We consider three settings:
\begin{enumerate}[leftmargin=*]
  \item \emph{Independent instances:} the agent solves each problem in a fresh context, without retaining state between instances.
  \item \emph{Online stream:} the agent receives problems one at a time from a problem family. The agent answers each problem and may retain a state (memory) between them.
  \item \emph{Compilation:} the agent receives a specification of the family and must produce a reusable solver that is frozen before evaluation.
\end{enumerate}
In the first two regimes, the agent produces answers to individual systems; in compilation, it produces a reusable solver artifact whose execution is evaluated on held-out problem instances. In this case, we distinguish the upfront cost of constructing the solver from its subsequent execution cost.

\section{Experiments and results}
\label{sec:experiments}

We evaluate Claude Code agents using Claude Opus 5 in the three regimes described above. In the independent-instance and online-stream settings, they have access to \texttt{msolve}~\citep{msolve2021}, SageMath~\citep{sagemath}, Singular~\citep{singular}, and Macaulay2~\citep{M2}.

\paragraph{Independent instances.} \label{para:independent_instances}
We first ask whether an agent can recognize special structure in individual systems and use it for faster computation (Figure~\ref{fig:teaser-monomial}). To this end, we consider three families with known efficient reformulations, described in Appendix~\ref{app:independent-generators}. The agent receives the equations without structural hints and solves each instance in a fresh context, without retaining any memory across instances. Table~\ref{tab:independent} compares direct \texttt{msolve}, a structural oracle using the known reformulation, and the agentic solver, across three size settings per family with 12 fresh systems per setting. Within the 600-second budget, the agent solves 103 of 108 systems, compared with 60 for direct \texttt{msolve} and 108 for the oracle. The agent solves every system completed by direct \texttt{msolve}, plus 43 additional systems on which the direct solver times out. However, among the 60 systems completed by both methods, the agent is faster on only 12, and its wall time remains much larger than the oracle's. These results show that agents can extend the range of problems solvable within a fixed time budget, but reasoning overhead can outweigh computational savings on easier instances. Additional diagnostics are given in Appendix~\ref{app:independent-diagnostics}.

\begin{table}[t]
\centering
\caption{Independent-instance setting: solved counts and median wall time per attempt (s) over 12 systems per configuration with a 600\,s budget. For eigenvalue systems, $m$ is the matrix size and $d$ the polynomial degree. \emph{Direct}: \texttt{msolve} on the input equations. \emph{Oracle}: structural solver using the supplied hidden structure. \emph{Agent}: Claude Opus 5 without structural hints. Times include unsuccessful attempts and exclude independent verification.  $^\dagger$All attempts timed out.}
\label{tab:independent}
\small
\setlength{\tabcolsep}{4pt}
\begin{tabular}{llrrrrrr}
\toprule
& & \multicolumn{2}{c}{Direct \texttt{msolve}} & \multicolumn{2}{c}{Oracle} & \multicolumn{2}{c}{Agent} \\
\cmidrule(lr){3-4}\cmidrule(lr){5-6}\cmidrule(lr){7-8}
Family & Size & solved & s & solved & s & solved & s \\
\midrule
Hidden monomial & $k=2$ & \textbf{12/12} & 2.2 & \textbf{12/12} & 0.046 & \textbf{12/12} & 229 \\
 & $k=3$ & \textbf{12/12} & 228 & \textbf{12/12} & 0.046 & \textbf{12/12} & 261 \\
 & $k=4$ & 0/12 & $600^\dagger$ & \textbf{12/12} & 0.046 & \textbf{12/12} & 230 \\
\midrule
Hidden triangular & $n=12$ & \textbf{12/12} & 9.3 & \textbf{12/12} & 0.035 & \textbf{12/12} & 127 \\
 & $n=16$ & 0/12 & $600^\dagger$ & \textbf{12/12} & 0.049 & 9/12 & 297 \\
 & $n=20$ & 0/12 & $600^\dagger$ & \textbf{12/12} & 0.052 & 10/12 & 286 \\
\midrule
Polynomial eigenvalue & $m=9,d=6$ & \textbf{12/12} & 9.0 & \textbf{12/12} & 0.063 & \textbf{12/12} & 145 \\
 & $m=10,d=6$ & \textbf{12/12} & 152 & \textbf{12/12} & 0.071 & \textbf{12/12} & 139 \\
 & $m=11,d=6$ & 0/12 & $600^\dagger$ & \textbf{12/12} & 0.081 & \textbf{12/12} & 121 \\
\bottomrule
\end{tabular}
\end{table}

\begin{figure}[t]
  \centering
  \includegraphics[width=\linewidth]{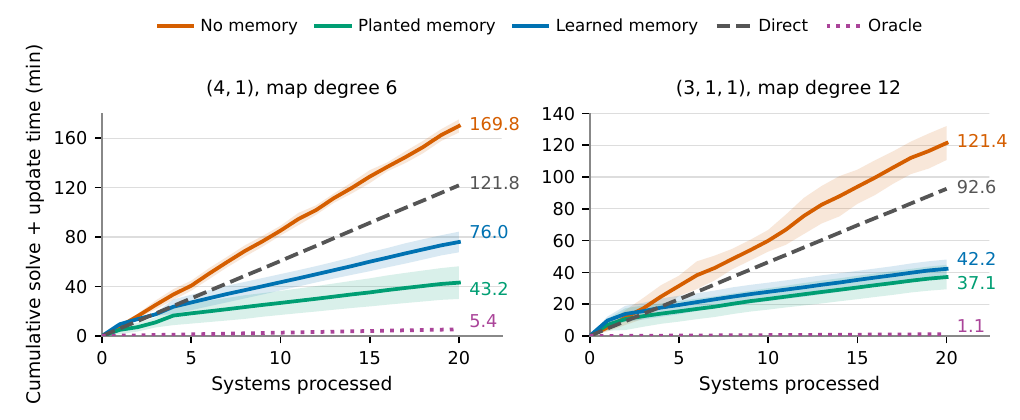}
  \caption{ Online streams of 20 quintic systems: $(4,1)$ with coefficient-map degree $6$ (left), and $(3,1,1)$ with degree $12$ (right). Agent curves show mean cumulative solve-plus-update time over three repetitions on identical inputs and order; shading is $\pm1$ sample standard deviation. Costs include unsuccessful attempts but exclude independent verification. Completion counts are in Table~\ref{tab:online-completion}.}
  \label{fig:online-stream}
\end{figure}

\paragraph{Online stream.}
We next investigate whether memory can reduce the cumulative cost of solving related systems. In our experiments, we use system families constructed by requiring a quintic with polynomially parameterized coefficients to have prescribed root multiplicities. Within each stream, the multiplicity pattern and coefficient-map degree are fixed, while the coefficient maps vary between instances (Appendix~\ref{app:multiple-root-family}). Figure~\ref{fig:online-stream} shows two streams of 20 systems: pattern $(4,1)$ with coefficient-map degree $6$ (left), and pattern $(3,1,1)$ with coefficient-map degree $12$ (right). Each solve has a 600\,s wall-time budget, with up to 180\,s per memory update. Cumulative costs include the time spent on unsuccessful attempts. For each pattern, a fixed ``multiple-root locus'' ideal describes the allowed coefficient vectors; substituting the coefficient maps yields a system with fewer unknowns, after which the eliminated coordinates can be recovered. Our Oracle baseline uses this approach with \texttt{msolve}. In addition to evaluating learned memory starting from an empty state, we also consider a ``planted memory'' that is initialized with the ideal for the family and recovery guidance, and which remains writable throughout the stream.

Without memory, the agent incurs greater cumulative cost than direct \texttt{msolve}.  Learned memory reduces cumulative cost relative to both the no-memory agent and direct \texttt{msolve}, but remains substantially slower than Oracle. Inspection of the saved memories and submitted reformulations shows that learned memory discovers a partial reduction within the first few systems and then repeatedly reuses it, without adopting the full multiple-root locus ideal. Planted memory uses the full reduction for $(4,1)$, whereas both memory conditions retain partial reductions for $(3,1,1)$ (Appendix~\ref{app:online-diagnostics}). For learned memory, per-instance cost falls rapidly at the start of both streams and subsequently levels off. The notes increasingly emphasize replaying the established pipeline, suggesting that memory can preserve a successful but suboptimal strategy.

\paragraph{Compilation.}
We next ask whether an agent can implement a fast reusable solver for a known problem family. We use an autoresearch loop as implemented by AutoHelix~\citep{autohelix2026}, a meta-harness that repeatedly asks an agent to modify a solver, evaluates candidate artifacts on a fixed development set, and retains verified improvements. The resulting solver is frozen before evaluation on held-out systems. We perform three compilation runs for the repeated root problem family with multiplicity pattern $(4,1)$, using 20 development systems with coefficient-map degrees $3$--$6$. We evaluate the resulting solvers on 60 held-out systems from the same family, with ten at each degree $1$--$6$.

\begin{figure}
    \centering
    \includegraphics[width=\linewidth]{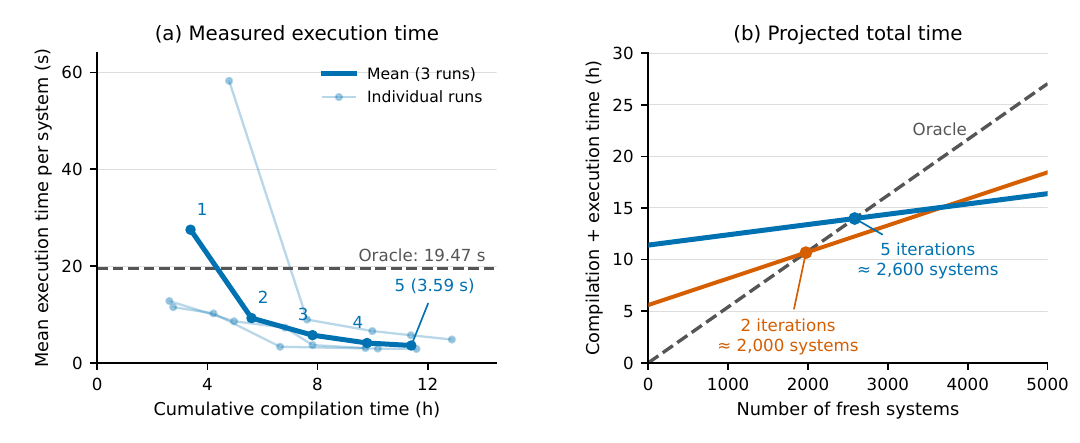}
    \caption{Compilation on the $(4,1)$ family. (a) Mean (bold) and individual (light) execution times across three runs, against cumulative compilation time; numbers denote accepted iterations. (b) Projected compilation-plus-execution time after two or five iterations, using mean compilation costs and execution times across runs.
    }
    \label{fig:compiled_solver_comparison}
\end{figure}

All three final solvers complete all 60 held-out systems, averaging $3.6$\,s per system versus Oracle's $19.5$\,s ($5.4\times$ faster). Two solvers recover the full coefficient-locus reformulation used by Oracle in the stream setting; the third retains a partial reduction. All three use \texttt{msolve} over finite fields together with their own code to recover exact RURs with rational coefficients. This speedup takes $11.4$ hours of compilation on average, with a projected break-even at $2{,}600$ systems.  Earlier stopping trades speed for a lower upfront cost: after two iterations, the solvers average $9.3$\,s per system with $5.6$ hours of compilation, breaking even at about $2{,}000$ systems. Both projections assume the same degree mixture as the held-out set
(Figure~\ref{fig:compiled_solver_comparison}).

\section{Discussion}
\label{sec:discussion}

We have proposed agentic computational algebra as a domain for studying how agents discover, memorize, and compile reusable computational structure. The setting  supplies infinitely many verifiable problems with large differences in cost between naive and structure-aware solution approaches. We described three evaluation regimes---independent instances, online streams, and known-family compilation---that expose different forms of agent experience.

On independent instances, agents extend the range of systems solvable within a fixed budget, but reasoning overhead often outweighs computational savings on already tractable problems. In the two quintic streams, memory reduces cumulative cost by preserving reusable reformulations. With learned memory, however, the agents settle on partial reductions and repeatedly replay them, remaining substantially slower than the structural oracle. These trajectories suggest that retaining a successful strategy does not necessarily lead to continued improvement. Compilation removes per-instance agent inference and produces solvers that outperform a baseline already supplied with the family's full reformulation. This gain requires substantial upfront computation, whose value depends on how many problems the solver handles and how much time it saves on each.

Future work could study when agents should continue exploring reformulations, refine their memory, or invest in a reusable solver, depending on the volume and diversity of subsequent problems. More broadly, evaluations could extend beyond zero-dimensional systems to semialgebraic problems or tasks requiring formalization from prose.

\bibliographystyle{plainnat}
\bibliography{main}

\appendix
\section{Output contract and verification}
\label{app:domain}
\label{app:contract}

We require solutions to be returned via a \emph{rational univariate representation} (RUR)~\citep{Rouillier1999-dg,msolve2021}. This is a standard exact representation of a finite solution set and is the format returned by \texttt{msolve}. For convenience, we recall the basic definition here.

\paragraph{Rational univariate representations.}
The first step is to choose a linear form
\[
  T(x_1,\ldots,x_n)=a_1x_1+\cdots+a_nx_n,\qquad a_i\in\QQ.
\]
For a solution $p \in \CC^n$, let $\tau_p=T(p)$. We call $T$ \emph{separating} if $\tau_p\neq\tau_q$ whenever $p\neq q$. Thus $\tau_p$ can be seen as a scalar label for the entire point $p$. Since there are only finitely many solutions, a generic choice of the coefficients $a_i$ is separating.

We collect these labels as the roots of a univariate polynomial
\[
  W(z)=\prod_p (z-\tau_p).
\]
Thus, choosing a root $\tau$ of $W$ selects one solution.
Now to recover the coordinates of the roots we define
\[
  g_i(z)=\sum_p x_i(p)\prod_{q\neq p}(z-\tau_q),
  \qquad v_i(z)=-g_i(z).
\]
Since $g_i(\tau_p)=x_i(p)W'(\tau_p)$, the solution
with label $\tau$ is given by

\begin{equation}
\label{eq:rur}
  W(\tau)=0,\qquad x_i=-v_i(\tau)/W'(\tau),
  \qquad i=1,\ldots,n.
\end{equation}

The polynomials $W$, $g_i$, and $v_i$ all have rational coefficients, since algebraic conjugation permutes the solutions and leaves the defining sum unchanged. Also $W'(\tau) \ne 0$ at every root, because labels are distinct. The polynomials $W, v_1,\ldots,v_n$ form an RUR of the solution set, with the solutions determined via~\eqref{eq:rur}.

\paragraph{Verification.}
The representation can be checked by manipulating only univariate polynomials with exact arithmetic. Consider the ring
\[
  A=\QQ[z]/(W).
\]
Since $W$ is squarefree, $W'$ is invertible in $A$. A submitted RUR therefore defines elements
\[
  \widehat x_i=-v_i(W')^{-1}\in A.
\]
For an input system $f_1=\cdots=f_m=0$, the verifier checks
\[
  f_j(\widehat x_1,\ldots,\widehat x_n)=0\quad\text{in }A
  \qquad (j=1,\ldots,m),
\]
where equality in $A$ means that the corresponding univariate polynomial has zero remainder modulo $W$. This ensures that every point represented
by~\eqref{eq:rur} satisfies the input equations.

To certify completeness (i.e., all solutions of the system are represented by the provided RUR), we use an upper bound $G$ on
the number of solutions, derived from the family construction.
We also check $T(\widehat x)=z$, ensuring that distinct roots
of $W$ represent distinct solutions. For a zero-dimensional
input system, the answer is complete when $\deg W=G$.

The identities above define an exact output contract.  In our experiments, for efficiency,  we verify these identities modulo four randomly chosen 60-bit primes for which the representation remains well defined. This avoids expanding the identities over \(\QQ\) and provides high-confidence probabilistic verification.

\paragraph{Lifted RURs.}
When an agent solves a reformulated system with \texttt{msolve}, the returned RUR describes the solutions in the new variables. Converting it to an RUR in the original variables can be expensive and require manipulating large coefficients. We therefore also accept a representation that augments the reformulated-system RUR with expressions for recovering the original variables.

For example, suppose the agent solves a system using variables
$y=\psi(x)$, with rational formulas $x=\varphi(y)$
recovering the original coordinates. If the reduced RUR uses the parameter $z=T(y)$, then
$\rho(x)=T(\psi(x))$ expresses this parameter in the original
coordinates. Writing $\widehat y$ for the coordinates
represented by the reduced RUR, the verifier checks
\[
  f_j(\varphi(\widehat y))=0,\qquad
  \rho(\varphi(\widehat y))=z
  \quad\text{in }A=\QQ[z]/(W),
\]
and requires all denominators to be invertible in $A$.
The first identity verifies the original equations while
the second ensures that distinct parameter values recover
distinct original solutions. The verifier also checks that every recovered solution is nonsingular and that the answer is complete ($\deg W=G$ as above).

\section{Benchmark families}
\label{app:families}

To evaluate agents, we construct polynomial systems with known structure and opportunities for more efficient solving. For independent instances,
we obscure a simpler problem through substitutions or an expanded
presentation, so that the agent must identify a useful reformulation.
For online streams and compilation, we construct related systems
for which part of the algebraic computation can be reused across
instances.

\subsection{Independent-instance families}
\label{app:independent-generators}

In the independent-instance setting, the agent receives equations without any description of the underlying structure. The Oracle solver implements a known simplification for each family. The families
used in our experiments are described below (Table~\ref{tab:independent-instance-details} lists the exact configurations).

\paragraph{Hidden monomial composition.}
We start with three cubic equations, replace their variables by
monomials, and expand. The resulting equations have high degree,
but recognizing the recurring monomials restores the cubic system.
For each instance, sample three dense cubics
$g_1,g_2,g_3\in\QQ[u_1,u_2,u_3]$, whose coefficients are drawn from
$\{-9,\ldots,-1,1,\ldots,9\}$. Generically, they have 27 solutions,
all with nonzero coordinates. Define the substitution by
\[
 E_k=
 \begin{pmatrix}1&0&0\\k&1&0\\k&k&1\end{pmatrix}
 \begin{pmatrix}1&k&k\\0&1&k\\0&0&1\end{pmatrix},
 \qquad
 u_i=\prod_{j=1}^3x_j^{(E_k)_{ij}}.
\]
The public equations fully expand $g_i(u_1,u_2,u_3)=0$.
Since $\det E_k=1$, the inverse matrix $B_k=E_k^{-1}$ has integer
entries and gives explicit recovery formulas
$x_j=\prod_i u_i^{(B_k)_{ji}}$.
These products may involve negative powers, so the substitution is
invertible where all coordinates are nonzero. For the evaluated
values $k>0$, all entries of $E_k$ are positive. If any $x_j$ were
zero, all three $u_i$ would be zero, contradicting the nonzero
constant terms of the cubics. Thus no public solutions are lost by
using the inverse map. Increasing $k$ raises the degree of the expanded system
while preserving the three-cubic reduced system and its generic
solution count.

\paragraph{Hidden triangular graph.}
We construct a triangular chain of equations:
\[
 y_1^9-a_1=0,\qquad
 y_i-s_i y_{i-1}^2-a_i=0\quad(i=2,\ldots,n),
\]
with $s_i\in\{-1,1\}$ and $a_i\in\{2,\ldots,11\}$.
Solving this system is straightforward, since
each of the nine roots of the first equation determines
all remaining coordinates through the recurrence.

We then substitute $\mathbf y=A\mathbf x$, where $A$ is
an integer matrix with determinant $\pm1$, and expand.
The agent receives the resulting equations in the $x$
coordinates without being given $A$. Recovering the hidden
coordinates allows the system to be solved by the recurrence,
followed by $\mathbf x=A^{-1}\mathbf y$.
Increasing the number of variables $n$ enlarges the system
while keeping the number of solutions fixed at nine.

\paragraph{Polynomial eigenvalue system.}
We start with a polynomial matrix
\[
 A(t)=A_0+A_1t+\cdots+A_dt^d,
 \qquad A_j\in\mathbb Z^{m\times m},
\]
and seek to solve
\[
 A(t)(v_1,\ldots,v_{m-1},1)^\mathsf T=0,
\]
in the variables $v_1,\ldots,v_{m-1},t$. In expanded form,
this gives $m$ polynomial equations in $m$ variables. However,
recognizing the matrix structure allows solving through the
univariate equation $\det A(t)=0$, which generically has
$md$ distinct roots, and each admits a unique vector with this normalization.

\begin{table}[t]
\centering
\small
\caption{Independent-instance configurations, with 12 fresh systems each.
Degree is the maximum total degree of a public equation; $v/e$ gives
variables/equations, and Reduced refers to the supplied structural
reformulation used by Oracle. Counts are the generic solution counts
used in qualification.}
\label{tab:independent-instance-details}
\begin{tabular}{@{}llrrrr@{}}
\toprule
Family & Setting & Public $v/e$ & Degree & Reduced $v/e$ & Solutions\\
\midrule
Hidden monomial & $k=2$ & $3/3$ & 51 & $3/3$ & 27\\
 & $k=3$ & $3/3$ & 102 & $3/3$ & 27\\
 & $k=4$ & $3/3$ & 171 & $3/3$ & 27\\
Hidden graph & $n=12$ & $12/12$ & 9 & $1/1$ & 9\\
 & $n=16$ & $16/16$ & 9 & $1/1$ & 9\\
 & $n=20$ & $20/20$ & 9 & $1/1$ & 9\\
Polynomial eigenvalue & $m=9,d=6$ & $9/9$ & 7 & $1/1$ & 54\\
 & $m=10,d=6$ & $10/10$ & 7 & $1/1$ & 60\\
 & $m=11,d=6$ & $11/11$ & 7 & $1/1$ & 66\\
\bottomrule
\end{tabular}
\end{table}

\subsection{Families with shared structure}
\label{app:shared-structure}
\label{app:family}
\label{app:multiple-root-family}

For the online stream and compilation experiments, we consider families
whose structure can be reused across instances. One way to do this is by
fixing polynomial equations and varying the polynomial maps
substituted into those equations.

Concretely, fix equations $F(\mathbf W,\mathbf U)=0$, where
$\mathbf W$ contains auxiliary variables and $\mathbf U$ contains
parameters. Each instance substitutes
$\mathbf U=\Phi_{d,\theta}(\mathbf Z)$, giving
\[
  F\bigl(\mathbf W,\Phi_{d,\theta}(\mathbf Z)\bigr)=0.
\]
Here $d$ is the degree of the polynomial map, and $\theta$
specifies its coefficients. The resulting system has unknowns
$\mathbf W$ and $\mathbf Z$.

Eliminating $\mathbf W$ from $F(\mathbf W,\mathbf U)$ gives constraints involving only
$\mathbf U$, which can be computed once for the family.
Substituting each instance's polynomial map then gives a system
in $\mathbf Z$ alone. Solving this smaller system can be much
cheaper than the original system, since there are fewer unknowns.
However, we must still then recover the
auxiliary variables and
check the original equations, since elimination can introduce extra
candidates.

In our experiments, we use this construction
to constrain coefficients of a univariate
polynomial to have prescribed root multiplicities.
For all patterns below, the coefficient maps $c_i(\mathbf Z)$
have degree at most $d$, with coefficients independently
sampled from $\{-5,\ldots,5\}$ on all monomials of total
degree at most $d$.

\paragraph{Quintics with pattern $(4,1)$.}
For pattern $(4,1)$, we require a quintic to have a root of
multiplicity four and one remaining root:

\begin{equation}
\label{eq:quintic}
 t^5+c_1(\mathbf Z)t^4+\cdots+c_5(\mathbf Z)
 =(t+a)^4(t+b),\qquad \mathbf Z=(Z_1,Z_2,Z_3).
\end{equation}

The unknowns are $\mathbf Z$ and the two factor coefficients $a,b$;
the variable $t$ is used only to express the factorization.
Equating coefficients gives the five-variable public system

\begin{equation}
\label{eq:quintic_coefficients}
\begin{aligned}
 c_1-4a-b&=0, & c_2-6a^2-4ab&=0,\\
 c_3-4a^3-6a^2b&=0, & c_4-a^4-4a^3b&=0,\\
 c_5-a^4b&=0.
\end{aligned}
\end{equation}

Here and below, $c_i$ abbreviates $c_i(\mathbf Z)$.
Only these maps change between instances at a fixed degree.
The set of coefficient vectors admitting this factorization has degree $8$~\citep{lee2015dualitymultiplerootloci}.
A generic degree-$d$ coefficient map in three variables
therefore gives $8d^3$ solutions ($1{,}728$ at $d=6$).

To obtain the shared reformulation, first treat the quintic
coefficients as independent variables $U_1,\ldots,U_5$ and
eliminate $a,b$. The resulting multiple-root locus ideal is

\begin{equation}
\label{eq:universal}
 I_{(4,1)}=\left\langle
 \begin{aligned}
  &20U_4-8U_1U_3+3U_2^2,\\
  &50U_5-6U_1U_4+U_2U_3,\\
  &20U_1U_5-8U_2U_4+3U_3^2
 \end{aligned}
 \right\rangle\subseteq\QQ[U_1,\ldots,U_5].
\end{equation}

This ideal depends only on the multiplicity pattern. Substituting
$U_i=c_i(\mathbf Z)$ gives the three-variable system solved by
Oracle. On the generic locus, the factor coefficients are recovered by
\[
 a=\frac{5c_3-c_1c_2}{10c_2-4c_1^2},\qquad b=c_1-4a.
\]
The verifier checks that the recovery denominator is invertible on
the submitted solution set. A simpler partial reduction substitutes
$b=c_1-4a$ into the remaining four equations and retains $a$,
leaving four variables rather than three.

\paragraph{General multiplicity patterns.}
More generally, let $\lambda=(\lambda_1,\ldots,\lambda_\ell)$ be a
partition of the univariate degree $n$. The multiple-root locus
consists of coefficient vectors admitting a factorization
\[
  t^n+U_1t^{n-1}+\cdots+U_n
  =\prod_{j=1}^{\ell}(t-R_j)^{\lambda_j}.
\]
Its generic points have $\ell$ distinct roots with multiplicities
$\lambda$; the locus also includes cases where roots
coincide~\citep{lee2015dualitymultiplerootloci}.

We group roots with the same multiplicity into a single factor,
so their ordering does not matter.
If $m_r$ parts of $\lambda$ equal $r$, introduce a monic polynomial
$Q_r(t)$ of degree $m_r$ and write
\[
  t^n+U_1t^{n-1}+\cdots+U_n=\prod_r Q_r(t)^r.
\]
The $\ell$ coefficients of these factors are the auxiliary
variables $\mathbf W$. Equating coefficients and replacing
$U_i$ by $c_i(\mathbf Z)$, in $k=n-\ell$ variables, gives $n$
equations in $n$ unknowns. Eliminating the factor coefficients
gives the reusable ideal for the chosen multiplicity pattern.

The degree of the coefficient locus is~\citep{lee2015dualitymultiplerootloci}
\[
  \delta_\lambda=
  \frac{\ell!\prod_{j=1}^{\ell}\lambda_j}{\prod_r m_r!}.
\]
For generic degree-$d$ coefficient maps, the number of isolated
solutions is $G=\delta_\lambda d^{n-\ell}$. This also bounds the
number of distinct isolated solutions for special coefficient maps:
each isolated zero of the square system persists with positive local
multiplicity under a sufficiently small generic perturbation, whose
total solution count is $G$. The benchmark uses this uniform bound
in its completeness check.

\paragraph{Quintics with pattern $(3,1,1)$.}
Here we require a triple root and two remaining roots.
Grouping the latter in a quadratic factor gives

\[
 t^5+c_1(\mathbf Z)t^4+\cdots+c_5(\mathbf Z)
 =(t+a)^3(t^2+bt+c),\qquad \mathbf Z=(Z_1,Z_2).
\]

Equating coefficients gives five equations in $Z_1,Z_2,a,b,c$:
\[
\begin{aligned}
 c_1&=3a+b, & c_2&=3a^2+3ab+c,\\
 c_3&=a^3+3a^2b+3ac, & c_4&=a^3b+3a^2c,\\
 c_5&=a^3c.
\end{aligned}
\]
The generic count is $9d^2$, or $1{,}296$ at $d=12$.
A partial reduction recovers $b=c_1-3a$ and
$c=c_2-3ac_1+6a^2$, leaving three equations in $Z_1,Z_2,a$.
An equivalent formulation imposes
$p(-a)=p'(-a)=p''(-a)=0$ for the quintic $p$.
Oracle eliminates all three factor coefficients
and solves the multiple-root locus ideal in $Z_1,Z_2$.
To keep the recovery expressions compact, it adds $a$ to the
reduced RUR and supplies short polynomial lifts for $b,c$.
The submitted reduced RUR therefore has three variables,
although the system solved by Oracle has only two.

\paragraph{Supplemental sextics with pattern $(2,2,2)$.}
The additional stream requires a sextic to be the square of a
cubic, giving three double roots generically:
\[
 t^6+c_1(\mathbf Z)t^5+\cdots+c_6(\mathbf Z)
 =(t^3+at^2+bt+c)^2,\qquad \mathbf Z=(Z_1,Z_2,Z_3).
\]
At map degree $6$, the generic count is again $8d^3=1{,}728$.
The Oracle eliminates the three cubic coefficients and recovers
them polynomially using
$a=c_1/2$, $b=(c_2-a^2)/2$, and $c=(c_3-2ab)/2$.

\section{Experimental protocols}
\label{app:protocols}

\subsection{Common conventions}

Agents use Claude Opus~5. In the independent-instance and stream
experiments, they have shell and filesystem access inside isolated
containers with \texttt{msolve}, SageMath, Singular, and Macaulay2
installed. Network access is limited to the model endpoint, and agents
cannot access the private generation, Oracle, or verification code.
Instructions describe the answer format and encourage exact
reformulations, compact recovery maps, and early submission for host
verification. The task prompts are
reproduced in Appendix~\ref{app:task-prompts}.

Measured solve time includes processing the input, performing algebraic
computations, and preparing the submission; agent time also includes
model inference.
Container startup and independent host verification are excluded.
Unsuccessful attempts, including timeouts, contribute their recorded
elapsed times. Independent-instance and stream experiments allocate
eight CPU cores and 32\,GiB per solve. Main compilation evaluations
allocate four cores and 12\,GB on a separate host. Timing comparisons
are made within each experiment.

\subsection{Independent instances}
\label{app:independent-protocol}

There are 108 inputs: 12 fresh systems at each of the nine
configurations in Table~\ref{tab:independent-instance-details}.
Each input receives one agent attempt in a fresh context with no
persistent state. All methods use the same inputs, a 600-second
solve budget, and the LiftedRUR contract.
Direct runs \texttt{msolve} on the original equations with eight
threads. Oracle solves the known monomial or determinant
reformulation with \texttt{msolve}; for the graph family, it constructs
the RUR by exact back-substitution without calling \texttt{msolve}.

Before running the agents, we checked that each input was well formed
and that Oracle returned a complete answer. Solver runtime was not
used to select inputs. The five unsuccessful agent attempts were two
graph timeouts and three graph submissions rejected for exceeding the
100{,}000-digit coefficient limit. All remain in the reported counts and medians.

\subsection{Online stream}
\label{app:online-protocol}

Each stream contains 20 fixed systems at a single coefficient-map
degree: $(4,1)$ at degree 6 and $(3,1,1)$ at degree 12 in the main
figure, and $(2,2,2)$ at degree 6 in the supplemental results.
We repeat each agent condition three times, using the same inputs in
the same order within each family. Direct and Oracle are evaluated
once per family. All methods use the LiftedRUR contract.
Direct invokes eight-thread \texttt{msolve} on the public equations.
Oracle uses the full coefficient-locus ideal. Its time includes
per-instance extraction, substitution, solving, and coordinate recovery,
but excludes construction of the shared ideal.

The no-memory condition resets context and files for each system.
Learned memory starts with an empty text file; planted memory starts
with the coefficient-locus ideal and recovery guidance.
In both memory conditions, the file remains writable, and each
repetition starts from its initial state. Only the text file carries over between systems;
conversation state and executable files are reset. Each solve has a
600-second cap, and each memory update has a separate 180-second cap.
After each attempt, the agent receives the host's verification verdict
and is asked to update its memory, including after unsuccessful attempts.
There is no update after the final system.
Reported costs include solve and update time. All conditions receive
instructions stating that the stream comes from a fixed family with
changing coefficients; no pattern or ideal is explicitly supplied to the
no-memory or learned-memory conditions. The public factor-variable
names are retained, including their multiplicity labels.

Before running any agents, we replaced one $(3,1,1)$ input because
Oracle did not certify completeness. We used the first candidate in
a fixed replacement list for which Oracle returned a complete answer,
keeping the other 19 inputs and their order. Selection depended only
on completeness.

\subsection{Known-family compilation}
\label{app:compilation-protocol}

The main experiment consists of three AutoHelix~\citep{autohelix2026}
compilation runs for the $(4,1)$ family. Each starts from the default
empty solver stub and receives the family specification, answer
contract, and the same 20 development systems with degrees
3--6. The coefficient-locus ideal is not supplied. Solvers may use
installed algebra backends and contain multiple files.
Hardcoding instance identifiers, coefficients, or precomputed answers
is prohibited.
The loop evaluates candidates on the development set and retains
verified improvements, with a 600\,s solve budget per development
system. Each run has a limit of five iterations or seven days;
all three complete five accepted iterations within this time.
Compilation cost is the total elapsed time, including development
evaluation and verification.

Frozen solvers are evaluated on the same 60 held-out
systems, ten at each degree 1--6, drawn independently of the
development pool, with a 1,600\,s solve budget per system.
All 15 accepted checkpoints solve all 60 systems;
held-out results do not influence which candidates the loop accepts.
Independent held-out verification is excluded from execution times.
All methods use the full RUR contract in the original variables.
The environment uses AutoHelix~0.1.1 and \texttt{msolve}~0.10.1.

Runs 2 and 3 use an updated training image with benchmark fixes and
larger output limits; the
\texttt{msolve} binary and \texttt{python-flint} version are unchanged.
Held-out checkpoints and Oracle are timed one at a time, with the
host otherwise idle. The projected payback uses the final mean
execution times and assumes that future inputs are equally distributed
across degrees 1--6, as in the held-out set. It extrapolates beyond
the 60-system test.

\paragraph{Oracle output and recovery.}
The $(4,1)$ stream and main compilation Oracles solve the same
coefficient-locus relations. The stream Oracle returns a lifted RUR
with symbolic recovery formulas, while main compilation requires a
full RUR expressing every original coordinate directly in the
univariate parameter.
The compilation Oracle solves the three-variable system with
\texttt{msolve}, then constructs the original-coordinate polynomials
using serial modular reconstruction and checks the original equations
internally. This work is included in its execution time.
This additional recovery work, along with differences in inputs,
hardware, and core allocations, limits comparisons between the two
Oracle timings.

\subsection{Task prompts}
\label{app:task-prompts}

The following task instructions are reproduced from the frozen run
files, with line breaks added for readability. The referenced public
guide supplies the LiftedRUR format and a small example.

\noindent\begin{minipage}{\linewidth}
\paragraph{Independent-instance solve prompt.}
All 108 independent instances use the same task instructions:
\begin{lstlisting}[basicstyle=\ttfamily\footnotesize,frame=single]
Solve the polynomial system in problem.ms exactly. You have at most 600 seconds of
wall-clock time. Minimize time to a complete, correct submission in submission.json.
Use only the shell and installed algebra tools. Before committing to a costly general
solve, look for exploitable algebraic structure or a simpler formulation. You may
solve directly or reformulate. Read /opt/benchmark/LIFTED_RUR.md for the public answer
format and a small format example. Once you have a mathematically justified
submission, finish promptly; the host verifies it, so avoid prolonged independent
audits. Do not inspect or reveal credentials. Write large solver outputs to files and
print only short summaries.
\end{lstlisting}
\end{minipage}

\paragraph{Stream solve prompt.}
This prompt is used for both memory conditions. The no-memory
condition omits the final instructions beginning ``Read
/work/MEMORY.md.''
\begin{lstlisting}[basicstyle=\ttfamily\footnotesize,frame=single]
Solve the polynomial system in problem.ms exactly. You have at most 600 seconds of
wall-clock time. Minimize time to a complete, correct submission in submission.json.
All problems in this stream come from one fixed generative family with changing
coefficients. Before committing to a costly general solve, look for reusable algebraic
structure or a simpler formulation. You may solve directly or reformulate. Use only
the shell and installed algebra tools. Read /opt/benchmark/LIFTED_RUR.md for the
public answer format and a small example. The container has 8 allocated CPU cores and
32 GiB of memory. For exact RUR output use `msolve -P 2 -t 8 -f input.ms -o
output.rur`. Msolve, SageMath, Macaulay2, Singular, and Python with python-flint are
available. Use `sage script.sage` or `sage script.py`; plain Python does not have
SymPy. Once you have a mathematically justified submission, finish promptly; the host
verifies it, so avoid prolonged independent audits. Do not inspect or reveal
credentials. Write large solver outputs to files and print only short summaries. Read
/work/MEMORY.md before solving and use any relevant reusable methods. Only the text in
that file can persist between problems; notes may include reusable code as text, but
executable files and the model conversation are not carried forward.
\end{lstlisting}

\paragraph{Memory-update prompt.}
The placeholder \texttt{<verdict>} is replaced by the current
case's host feedback.
\begin{lstlisting}[basicstyle=\ttfamily\footnotesize,frame=single]
You have at most 180 seconds of wall-clock time to update /work/MEMORY.md. Read its
existing contents and preserve useful mathematics and implementation lessons from the
current problem. Only this text persists; reusable code may be recorded as text. Write
useful notes promptly. Small symbolic computations are appropriate, but do not launch
another full-instance solve or a prolonged audit. Record only observed outcomes;
distinguish verified results from untested ideas. Do not store instance-specific
answers or credentials. Use only the shell. Do not write a submission. Verified host
feedback: <verdict>
\end{lstlisting}

\section{Diagnostic details and additional results}
\label{app:additional-results}

\subsection{Independent-instance reformulations}
\label{app:independent-diagnostics}

The following examples show how agents reformulated the first
instance in each of three hard configurations: hidden monomial
composition with
$k=4$, polynomial eigenvalue systems with $m=11,d=6$, and hidden
triangular graphs with $n=16$. The agent returned all solutions
in each example, while Direct timed out. In addition to the usual
verification, we checked these submissions exactly over $\QQ$.

\paragraph{Hidden monomial.}
For the $k=4$ example illustrated in Figure~\ref{fig:teaser-monomial},
the agent recognized that the expanded equations are cubic
polynomials in three recurring monomials. It introduced the coordinates
\[
 y_1=x_1x_2^4x_3^4,\quad
 y_2=x_1^4x_2^{17}x_3^{20},\quad
 y_3=x_1^4x_2^{20}x_3^{33},
\]
reducing the degree-171 public equations to three dense cubics.
The inverse monomial substitution recovers the original coordinates,
as described in Appendix~\ref{app:independent-generators}.
The agent returned a degree-27 RUR for the cubic system with
parameter $y_3$, together with the recovery formulas. \texttt{msolve} solved the reduced system in under 0.1 seconds; total agent time was 213.5 seconds.

\paragraph{Polynomial eigenvalue.}
For the $11\times11$ degree-six matrix system, the agent submitted
the degree-66 equation $\det A(t)=0$ and used ratios of matrix
cofactors to recover the ten normalized eigenvector coordinates.
The ratios share a denominator that is invertible modulo the
eliminant. Both the reduced and lifted representations use $t$
as the parameter. The attempt took 116.5 seconds in total.

\paragraph{Agent response excerpt.}
The agent's final response for the polynomial eigenvalue example
includes the following explanation. Bracketed ellipses mark
omitted text.
\begin{lstlisting}[basicstyle=\ttfamily\footnotesize,frame=single]
**Structure exploited:** every equation is *linear* in `v1..v10` with degree-6
coefficients in `t`. Writing the system as `A(t)·(v1,…,v10,1)ᵀ = 0` with `A` an 11×11
polynomial matrix:

[...]

**Submission:** reduced system is the univariate `t`-equation `det A(t) = 0`;
`artifact` is verbatim `msolve -P 2` output on it (eliminant proportional to `det A`,
`q' = E'`, empty numerator list, as expected for one variable); `parameter_recovery`
is `t`.
\end{lstlisting}

\paragraph{Hidden graph.}
The agent recovered the triangular coordinates, including
$y_1=3x_1+x_4+x_7$, and submitted the 16-variable chain with its
variables ordered so that \texttt{msolve} used $y_1$ as the parameter.
The eliminant is $T^9-6$, and the submitted inverse linear map
includes $x_1=y_1-y_7$. The agent thus exploited the triangular
structure while still submitting a 16-variable system. The attempt
took 444.8 seconds in total.

\subsection{Online-stream behavior}
\label{app:online-diagnostics}

Table~\ref{tab:online-completion} gives completion counts and total
costs for the two main streams and an additional sextic family.
The shaded bands in Figure~\ref{fig:online-stream} show the sample
standard deviation at each stream position across three repetitions
on the same inputs.

\begin{table}[t]
\centering
\small
\caption{Online-stream completion and total cost. Entries are mean minutes
$\pm$ one sample standard deviation, followed by complete attempts in
parentheses. Agent counts aggregate three 20-system repetitions; baseline
counts are from one evaluation. Failed attempts and memory updates are
included; independent verification is excluded.}
\label{tab:online-completion}
\begin{tabular}{@{}lccc@{}}
\toprule
Method & $(4,1),\ d=6$ & $(3,1,1),\ d=12$ & $(2,2,2),\ d=6$\\
\midrule
Direct & $121.82\ (20/20)$ & $92.58\ (20/20)$ & $115.41\ (20/20)$\\
Oracle & $5.40\ (20/20)$ & $1.12\ (20/20)$ & $15.10\ (20/20)$\\
No memory & $169.79\pm5.30\ (49/60)$ & $121.37\pm10.68\ (58/60)$ & $175.21\pm1.16\ (51/60)$\\
Planted memory & $43.15\pm13.21\ (58/60)$ & $37.07\pm7.68\ (59/60)$ & $57.80\pm10.91\ (57/60)$\\
Learned memory & $76.00\pm8.34\ (59/60)$ & $42.24\pm5.85\ (59/60)$ & $147.03\pm9.94\ (57/60)$\\
\bottomrule
\end{tabular}
\end{table}

\paragraph{Retained reductions.}
For $(4,1)$, all 59 successful learned-memory submissions eliminate
only $b$, using $b=c_1-4a$. They retain $a$ alongside
$Z_1,Z_2,Z_3$, leaving a four-variable system. This is a partial
reduction: the full coefficient-locus ideal eliminates both $a$
and $b$, leaving only the three variables $Z_1,Z_2,Z_3$.
All 58 successful planted-memory submissions use this full
three-variable reduction. In each learned-memory repetition,
the partial reduction is already present in the first successful
solve. Later notes mainly add implementation guidance, and costs
show no clear sustained decrease after the first few systems,
even though the memory continues to change. The final system
has no memory update, which partly explains its lower cost.

For $(3,1,1)$, all successful submissions in both memory conditions
use three-variable partial reductions that retain $a$.
These reformulations use polynomial division or equivalent
triple-root derivative conditions and were checked exactly against
the original equations. The planted memory includes both the full
ideal and the partial reduction; all successful planted-memory
submissions use the latter. Oracle solves in two variables before
recovering the factor coordinates. Supplying structural knowledge therefore does not
ensure that the agent uses the full elimination.

\paragraph{Saved-memory excerpt.}
After the sixth system in the first learned-memory $(4,1)$
repetition, the memory recommends replaying the partial reduction
and discourages further exploration:
\begin{lstlisting}[basicstyle=\ttfamily\footnotesize,frame=single]
I.e. following these notes verbatim costs ~7 tool calls and   ~1/4 of the 600 s
budget; do NOT explore alternatives, there is no payoff.

[...]

For (4,1): substitute `u1_1 = Q1 - 4*u4_1`. => reduced system: 4 vars (z1,z2,z3,u4_1),
4 eqs of degrees 7,8,9,10, ~160 terms each, solution count (deg of RUR) 1728.
\end{lstlisting}

\paragraph{Additional sextic stream.}
The $(2,2,2)$ stream follows the same protocol and has the same
generic solution count of 1,728 as the $(4,1)$ stream. Learned memory
reduces cost relative to no memory, but remains slower than Direct
in all three repetitions; planted memory is faster than Direct in
all three. In the first two learned-memory repetitions, every
successful submission retains the original six-variable system,
even though the notes recognize that the sextic is a square of a
cubic. Here, too, memory preserves a workable procedure without
using the available elimination.

\subsection{Compilation trajectory}
\label{app:compilation-trace}

All three final solvers outperform the structural Oracle.
Their mean execution times are 3.06, 2.88, and 4.82 seconds;
compilation takes 9.74, 11.59, and 12.88 hours, respectively.
At its first accepted checkpoint, the third solver takes 58.26
seconds per system, compared with Oracle's 19.47 seconds.

\paragraph{Compiled solving pipeline.}
Two final solvers use the three-variable coefficient-locus
reformulation in Equation~\eqref{eq:universal}; the third eliminates
only the simple-root variable, retaining four variables. In their
main solving paths, they first call \texttt{msolve} to compute a
parametrization modulo a prime $p$. They then use Newton lifting
to refine it modulo higher powers of the same prime: under the
required nonsingularity conditions, each step can increase the
modulus from $p^k$ to $p^{2k}$. A coefficient's value modulo $p$
is not enough to determine its rational value. Increasing the modulus
provides more information for reconstructing coefficients with large
numerators and denominators.

The solvers combine these lifted results from several starting
primes using the Chinese remainder theorem, then recover rational
coefficients using lattice-based reconstruction with shared
denominators. They reconstruct the eliminated factor coordinates
in the same way, giving a full RUR in the original variables.
The solvers check the original equations modulo additional primes and use fallback
methods when needed. The benchmark verifies the returned RUR
independently. \texttt{msolve} already uses modular computations and rational reconstruction to produce exact RURs. The compiled solvers use its finite-field mode together with their own Newton lifting, reconstruction, and coordinate recovery; the reported speedups compare these complete pipelines.

\paragraph{Compilation without general algebra backends.}
In a separate experiment, we ask the agent to implement the
algebraic solving steps itself, without calling \texttt{msolve},
Singular, or any other polynomial-system or Gr\"obner-basis backend.
We supply the $(4,1)$ coefficient-locus ideal and fix the
coefficient-map degree at 6, so the task focuses on implementing
a fast solver for a known reformulation. GMP/FLINT arithmetic,
univariate polynomial operations, and linear algebra are allowed.
The resulting solver uses Python and a C kernel
with FLINT arithmetic to perform modular Gr\"obner-basis reductions,
quotient-ring linear algebra, and rational reconstruction.

Using four development systems, AutoHelix rejects iterations 1--2
and accepts iterations 3--6. We freeze the iteration-6 solver and
evaluate it on 32 fresh degree-six systems with eight cores,
32\,GiB, a 180-second solve cap, and the same LiftedRUR contract
as Oracle. Both solve all 32 systems. The compiled solver takes
8.28 seconds on average, with a median of 3.06 seconds, compared
with Oracle's mean of 16.07 seconds: a $1.94\times$ mean speedup.
Three cases take approximately 59 seconds each, contributing to
the difference between mean and median. After correcting an
arithmetic-validation check, the solver again solves all 32 inputs,
with a mean time of 8.30 seconds.

Total compilation time is 4.31 hours, including a final rejected
iteration that leaves the solver unchanged. Using the measured
mean execution times, this cost would be recovered after approximately
2,000 systems, extrapolating beyond the test pool. The supplied
information, degree range, resources, and output representation
differ from the main compilation study, so the timings are not
directly comparable.

\section{Author use of AI tools}

AI agents are part of the experimental methodology and are documented
in the corresponding experimental sections. Separately, the authors
used AI-based coding and writing assistants to help implement and
debug experimental infrastructure and to assist with drafting,
editing, and polishing this manuscript. The authors reviewed the
resulting artifacts and take full responsibility for the correctness,
originality, citations, and final content of the work.

\end{document}